\documentclass[preprint,12pt]{elsarticle}

\usepackage{graphicx}
\usepackage{amssymb}
\usepackage{amsmath}
\usepackage{lineno}

\usepackage[font=footnotesize,labelfont=bf]{caption}
\usepackage[font=footnotesize,labelfont=bf]{subcaption}
\usepackage{url}

\journal{Remote Sensing of Environment}

\begin{document}

\begin{frontmatter}


\title{Mining and Tailings Dam Detection In Satellite Imagery Using Deep Learning}



%%\author{Remis Balaniuk $^{1}$*\orcidA{}, Olga Isupova $^{2}$  and Steven Reece $^{3}$}

\author{Remis Balaniuk $^{1}$, Olga Isupova $^{2}$  and Steven Reece $^{3}$}

% Authors, for metadata in PDF
%\authorNames{Remis Balaniuk, Olga Isupova and Steven Reece}

% Affiliations / Addresses (Add [1] after \address if there is only one affiliation.)
\address{%
$^{1}$ \quad Universidade Católica de Brasília and Tribunal de Contas da União, Brasília, Brazil; remisb@tcu.gov.br\\
$^{2}$ \quad Dept. Computer Science, University of Bath, Bath, UK; oi260@bath.ac.uk\\
$^{3}$ \quad Dept. Engineering Science, Oxford University, Oxford, UK; reece@robots.ox.ac.uk
}

\begin{abstract}
This work explores the combination of free cloud computing, free open-source software, and deep learning methods to analyse a real, large-scale problem: the automatic country-wide identification and classification of surface mines and mining tailings dams in Brazil. Locations of officially registered mines and dams were obtained from the Brazilian government open data resource. Multispectral Sentinel-2 satellite imagery, obtained and processed at the Google Earth Engine platform, was used to train and test deep neural networks using the TensorFlow 2 API and Google Colab platform. Fully Convolutional Neural Networks were used in an innovative way, to search for unregistered ore mines and tailing dams in large areas of the Brazilian territory.  The efficacy of the approach is demonstrated by the discovery of 263 mines that do not have an official mining concession. This exploratory work highlights the potential of a set of new technologies, freely available, for the construction of low cost data science tools that have high social impact. At the same time, it discusses and seeks to suggest practical solutions for the complex and serious problem of illegal mining and the proliferation of tailings dams, which pose high risks to the population and the environment, especially in developing countries. Code is made publicly available at: https://github.com/remis/mining-discovery-with-deep-learning.

\end{abstract}


\end{frontmatter}

%\linenumbers

\section{Introduction}
\label{S:1}

About 3.5 billion people live in countries rich in oil, gas or minerals~\footnote{https://www.worldbank.org/en/topic/extractiveindustries}. Natural resources have the potential to drive growth,  progress and poverty reduction in developing countries.  Despite its economic importance, mining is associated with major risks, both related to accidents and environmental damage.  Tailings dams are a by-product of mining, storing toxic waste liquid and solids.  Failure of tailings dams can be catastrophic, rapidly releasing large amounts of water and solid material, potentially causing large loss of life and huge damages to the environment and property.  On 25 January 2019 a tailings dam at the Córrego do Feijão iron ore mine in Brumadinho, Brazil, suffered a catastrophic slope failure, followed by a mudflow that killed at least 248 people. Three years and two months earlier, on November 5, 2015, Samarco's tailings dam broke down, destroying two villages, killing 19 people and seriously polluting 650 km of the Rio Doce River. Unfortunately, tailings dam failures are not rare events. In fact, there have been 46 failures world wide in the last 20 years and the number has steadily increased~\citep{ARMSTRONG2019101412}.  As well as the lives lost, the average financial cost of one very serious failure (i.e. a catastrophic dam failure that releases more than 1 million cubic meters of tailings) to loss of property and the economy has been estimated to be \$543 million \citep{bowker}.  The risks, the challenges for long-term containment and the relatively poor safety record revealed by the number of failures of tailings dams, have led to an increasing awareness of the need for enhanced safety provisions~\citep{icold2001tailings}.  

To control and prevent illegal or hazardous mining this paper offers a scalable and cost effective method for the wide area detection and impact risk assessment of tailings dams and mines.  The approach uses freely available multispectral geosatellite imagery, free cloud-based computational resources and recent advances in machine learning.

Most developed countries adopt frameworks for mining regulation that attempt to mitigate mining disasters whilst respecting community needs, the environment and keeping hazard risks under control~\citep{Cameron2017}.  Government-approved permits are usually required for all new and ongoing mining operations, including exploration activities. This permit process can ensure that environmental standards are maintained from the beginning to the end of mining and metal production operations. New mine developments are required to have operation and closure plans that define how a specific site will be reclaimed upon cessation of mining. This attention to the full life-cycle of a mining site is critical to preventing negative environmental impacts~\citep{miningusa}. 

Unfortunately, many developing countries still face a myriad of challenges, such as weak governance, lack of transparency and lack of accountability in extractive industries, preventing inclusive growth, while threatening both people and the environment.  Federal governments' lack of governance opens the door to informal and illegal mining and the fraudulent or incomplete reporting of activities by mining companies, causing major tax collection losses \citep{tcu} but also increasing risks to the environment and the population.
 
There is no complete inventory of active tailings impoundments around the world. The lack of any comprehensive tailings dam database has prevented meaningful analysis of the technical failures that could help prevent future incidents. The records are very incomplete on crucial data elements: design height of dam; design footprint; construction type (upstream, downstream, center line); age; design life; construction status; ownership status; capacity; release volume; runout etc. Risk management and early-warning of failure, like accelerating slope displacement, are essential to support decision making, but require frequent high quality data. 

Many illegal or informal mining operations take place in and around forests, sometimes overlapping with areas of high conservation value and high watershed stress that are home to vast amounts of biodiversity~\citep{congo,miningamazonia}.  These mining activities are often
difficult to detect from the ground.  However, satellite or air-borne sensing coupled with Geographic  Information  Systems  (GIS) can  be  used  to  assess  where  hazardous  natural or man-made phenomena  are occurring or are likely  to  occur over a wide area. This  information  can  feed  into  risk  assessments  or illegal activities detection, along  with  information  on  natural  resources, population and infrastructure. It can be used by governments and communities in order to prevent or stop dangerous or illegal activities but also to design mitigation strategies to reduce vulnerabilities to acceptable levels.

%Mapping hazards and potential  sources  of  disaster  using  GIS  provides  essential  data  for  disaster  risk  reduction  plans by  %allowing  governments and communities to  link  data  using  a  geographical  dimension.  Using  these  links, auditors and public authorities %can  discover  and  create  information  which can  be  translated  into  action  or  policies~\citep{ISSAI5510}. 

Unfortunately, the use of GIS technologies in complex and country-wide scale analysis is not currently available to all. It requires access to huge remote sensing imagery datasets, robust computation infrastructure, expensive software tools and teams of dedicated experts to interpret the vast amounts of imagery. Governments of developing countries are not always prepared, or do not have the financial resources, to deploy such a sophisticated infrastructure.

\section{The Machine Learning for Social Good Revolution}
New technological trends may help to reduce both the need for significant in-country technical expertise and computation infrastructure. Free cloud computing resources, free open-source software tools and deep learning methods can be used to build inexpensive and powerful data analysis tools. Easy access to these powerful technologies, coupled with the support that can easily be obtained from numerous online technical forums, tutorials and online training opens new frontiers for their unrestricted use by governments, entities and communities committed to the social good.  Anyone can access the latest free Earth observation imagery, from optical and radar satellite imagery to weather data and digital elevation maps. 40 years’ worth of free satellite images can be obtained from USGS-NASA Landsat \footnote{https://earthexplorer.usgs.gov/} or ESA's Copernicus Open Access Hub \footnote{https://scihub.copernicus.eu/}. Google Earth Engine \footnote{https://earthengine.google.com/} can be used at no cost for search, preview, and preprocessing of these large free GIS image sets.  

Remote-sensing image analysis in combination with machine learning methods have seen a massive rise in popularity over the past few years. Machine learning has been applied to tasks including image fusion, image registration, scene classification, object detection, land use and land cover (LULC) classification, segmentation, and object-based image analysis (OBIA) \citep{MA2019166}.  We will demonstrate that scene classification and image segmentation can be used to build comprehensive tailings dam databases, including their footprints, heights, volumes, changes over time and contained waste profile. These databases can be used for risk assessment and monitoring by regulatory agencies and local communities.  To explore the combination of free cloud computing, free software, and deep learning in analyzing a real large-scale problem, this paper demonstrates an application in automatic country-wide identification of surface mines and mining tailings dams in Brazil, followed by the classification of their potential environmental impact.  The approach
uses open-source and freely available software including Python \citep{python} and TensorFlow \citep{tensorflow} that can be run for free on Google cloud
infrastructure Colaboratory (or Colab)\footnote{https://colab.research.google.com/notebooks/welcome.ipynb}.  Our paper is accompanied by open-source Python code and data that can be adapted to other Earth observation (EO) applications by
government departments and their supervisory bodies and also by local communities and non-governmental organisations.   
Although this study focuses on mining activity in Brazil, the proposed approach is applicable to any region where satellite imagery is available. 

Section~\ref{sec:case-study} describes the environmental and social impact of illegal mining and motivates a Brazil case-study.  After presenting related work in Section~\ref{sec:relatedWork}, we describe the data processing methodology, specifically our convolutional neural network approach, and its application to mine detection and tailings dam environmental risk assessment in Section~\ref{sec:method}.
The efficacy of our approach is demonstrated on the Brazil case-study in Section~\ref{sec:results}.  We conclude in Section~\ref{sec:conclusions}.

\section{Mining Processes and the Brazil Case Study}
\label{sec:case-study}
\begin{figure}[h]
\centering
\includegraphics[width=0.8\columnwidth]{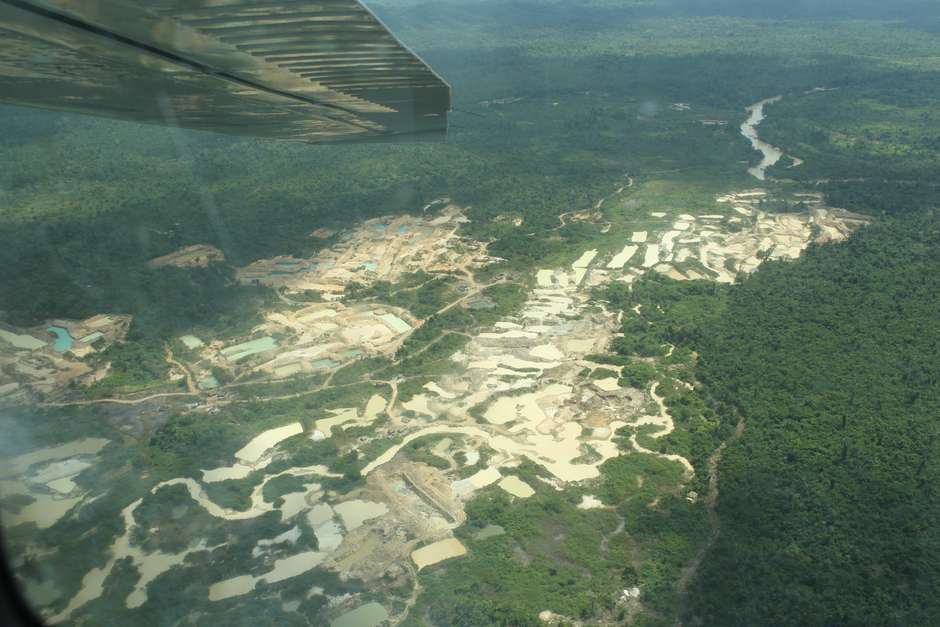}
\caption{Illegal mining fronts in Kayapó Indigenous Land, Pará . Photo: Ibama / BBC News Brasil.}
\label{fig:kayapo}
\end{figure}

Surface mining accounts for two thirds of the world’s metals used in industrial and technological applications and produces geochemically abundant metals such as aluminum, iron, magnesium, coal, manganese, and titanium, and also geochemically scarce metals, which include copper, lead, zinc, gold, and silver. Large-scale quarrying and open-cast mining is also used for stone, sand, gravel, crushed stone and phosphates extraction. Surface mining usually uses heavy equipment, such as earthmovers, to remove overburden, followed by large machines, such as dragline or bucket wheel excavators, to extract the mineral. Large mining companies typically use appropriate methods and equipment and assume responsibility for environmental damage or impacts. Illegal mining, by contrast, uses makeshift methods and never takes responsibility for the damage it causes. In some rainforest territories illegal miners use ferries to search the riverbed for precious metals. In others, as well as the rafts, there are almost industrial scale mining fronts, where backhoes and dredgers dig craters in the woods. This is the case for many of the mines in the Tapajós region of Pará, Brazil, where satellite and aerial images show large open "scars" in the forest, even on protected indigenous land (see, for example, Figure \ref{fig:kayapo}). 

Metal mining uses a huge amount of water in the crushing and ore separation process and leave behind residual rocks that have been reduced to the consistency of gravel or sand, called tailings.  Tailings are a mixture of particles suspended in liquid, usually toxic and potentially radioactive and can be deposited in large ponds. These ponds, called "tailings dams",  are enormous lake-like structures, square-kilometres in size,  and are held in place by earth-filled dams. The dams in mountainous valleys can reach as high as 300 metres.  These impoundments are designed for permanent containment and are some of the largest man-made structures on Earth.  Advances in mining technology have made it possible to exploit lower grade deposits despite decreasing commodity prices, creating more waste and putting more pressure on tailings facilities \citep{ARMSTRONG2019101412}. 

According to Brazilian legislation, those interested in exploring the subsoil for the extraction of ores must obtain a mineral exploitation concession through the mining concession regime, aimed at the extraction, processing and commercialization of the mineral asset. Before obtaining this concession, he must follow a series of steps, starting with the request for a research authorization  \citep{lei9406}. 

The Brazilian National Mining Agency (ANM) is responsible for supervising the mining activity in Brazil, from the authorization of mining to supervision of the enterprise, including the collection of taxes from mineral exploration and the inspection of tailings dams.  Unfortunately, ANM suffers from a poor administrative structure with insufficient staff numbers, a lack of training, technological deficiencies and an annual budget that has been cut by successive governments  \citep{mpmg}. In 2019 the agency had only 35 inspectors to inspect the 790 ore dams officially declared across the country\footnote{https://brasil.estadao.com.br/noticias/geral,pais-tem-apenas-35-fiscais-de-barragem-de-mineracao,70002699885}.  

ANM's work on controversial projects, including mining in conservation areas and indigenous lands, is also hampered by legal loopholes.
There is no consensus in the normative scope for mineral research in protected areas. The Brazilian federal constitution provides for research and mining in indigenous lands, but it requires that specific regulations be made by the National Congress, after hearing from the affected communities \citep{salesdissertacao}.
As this regulation does not yet exist, instead of rejecting proposals from mining companies, ANM can only attempt to delay them, arguing that they wait for the creation of an appropriate law.  Meanwhile, mining companies, some armed with provisional authorizations from the courts, continue their enterprises with impunity.

According to a WWF-Brazil report \citep{wwf}, there were in 2018 5,675 requests for mining concessions awaiting the approval of the ANM that overlap
totally or partially with protected areas within the scope of the Legal Amazon. This demonstrates the magnitude of the threat to these protect areas and the pressure to remove restrictions on mining.

\section{Related Work} \label{sec:relatedWork}

Remote sensing imagery has been used for surface mine monitoring for several decades \cite{Legg1990}. Recent increased availability and diversity of satellite image sets has sparked a wider range of studies to monitor environmental impact of mining activities using visual or computer-assisted interpretation with automated pattern recognition techniques \citep{YU2018675,RedondoVega2017,Petropoulos2013,LaJeunesseConnette2016}.   This work is, in part, driven by the lack of databases containing geographic coordinates, characteristics and attributes of mines.  These characteristics of mines are important for environmental management and monitoring. The United States Geological Survey (USGS) provides the Major Mineral Deposits of the World (MMDW) database\footnote{https://mrdata.usgs.gov/major-deposits/} containing locations of known surface mining areas around the world. \citet{YU2018675} used this database to investigate the effectiveness of remote sensing datasets to identify and map land cover changes related to surface mining activities, but reported a lack of precision related to geographic location, geometry and spatial extent of deposits. \citet{Soulard2015} employed semi‐automated procedures to map country wide mining footprint and change detection in US using mine seed points derived from official surveys and datasets.  However, to the best of our knowledge, no study has been done on automated discovery and classification of unregistered surface mining sites and tailings dams on wide areas.

Space-borne sensors with a wide range of spatiotemporal, radiometric and spectral resolutions have become valuable data sources for large-scale Earth observation applications. However, data availability from multispectral and hyperspatial sensors introduces new challenges in data mining, processing, backup and retrieval. Complexity and volume of these datasets require advanced data processing, flexible interfaces and computational power. Cloud-based platforms such as Google Earth and powerful open source software libraries are creating new opportunities and bringing in new users and new applications of remote sensing in environmental sciences and natural resources management. For a survey on processing  remote sensing data in cloud computing environments refer to \citep{Sugumaran}. A special issue of the Remote Sensing of Environment Journal was dedicated to the remote sensing of land change science with Goggle Earth Engine  \citep{remotesensinggee}. 

Machine learning methods have gained strength with these new resources. Deep learning methods have obtained state-of-the-art results in many Earth observation applications, such as image classification \citep{penatti2015deep}, semantic segmentation \citep{volpi2016dense}, phenological studies \citep{nogueira2019spatio}, poverty mapping \citep{xie2015transfer} and detection of pivot irrigation systems with very high spatial and temporal resolution imagery \citep{pivotirrigation}. An extensive survey on deep learning-driven remote sensing image scene understanding can be found in \citep{gu}.  In deep learning, convolutional neural networks (CNNs) play a central role for processing visual-related problems \citep{Hinton2006}. CNNs have been successfully employed, for instance, to classify hyperspectral images directly in spectral domain \citep{Hu2015}, pixelwise classification of satellite imagery \citep{Maggiori2017} and detection of informal settlements in very high  resolution (VHR) satellite  images \citep{Persello2017}. A special issue of the Remote Sensing Journal was dedicated to the applications of CNNs in remote sensing \citep{remotesensing}.

\section{Materials and Methods}  \label{sec:method}

\subsection{Data acquisition} \label{sec:data}

Deep learning methods require rich training datasets to fit, evaluate and test the neural network models.   Preparation of these datasets requires a considerable number of good quality images containing the targets of interest. For applications in remote sensing it is imperative to obtain a representative collection of satellite imagery and address several corresponding technical issues such as cloud cover, select appropriate spectral bands and undertake image registration. For supervised learning it is also paramount to have trusted sources of ground truth labelled imagery.   In this work we used, as a starting point, two official georeferenced databases released by the Brazilian federal government. The first is a catalog of 612 officially reported tailings dams in Brazil, made available by the Brazilian National Mining Agency on its website \footnote{http://www.anm.gov.br/assuntos/barragens/pasta-cadastro-nacional-de-barragens-de-mineracao/classificacao-oficial-anm}. This database contains some information about each dam including its owner and spatial coordinates of the dam's location. There is also some technical information including the main ore mined, height, volume, construction method, risk category, and potential damage that could arise if the dam were to fail. The second database is a catalog of 268 legally operating ore mines in Brazil obtained from the same agency, but is not publicly available. 

Unfortunately, both of these databases have quality issues. A significant number of coordinates point to locations near the mines or tailings dams but not exactly to them, or they are completely off target. A visual inspection of each reference using Google Earth images was required to address this problem and correct the target coordinates. \cite{ferreira2020brazildam} used the same databases to create the {\it BrazilDAM}, a  public dataset based on Sentinel-2 and Landsat-8 satellite images covering the tailings dams cataloged by the ANM.

Our machine learning classifiers require samples of all target classes to train them.  This includes the {\it background} class which are objects other than dams and mines. Thus, it was also necessary to create a database with coordinates of locations where no mines or tailings dams were present. A careful choice of representative locations was necessary in order to capture the diversity of the background class such as cities, construction sites and natural lakes. Consequently, two databases were created: one containing 1397 coordinates pointing to mines or tailings dams and another containing 1463 background locations.

As our goal was to implement an approach using only freely available resources that can be used to identify mines and dams in any region of the Brazilian territory, we chose to use only open access satellite imagery. The main providers are NASA and the European Space Agency (ESA), offering the Landsat Mission datasets\footnote{https://landsat.gsfc.nasa.gov/data/}  and the Copernicus-Sentinel program datasets\footnote{https://sentinel.esa.int/web/sentinel/sentinel-data-access} respectively. We chose to use the Sentinel images, given their better resolution and sampling frequencies. The Sentinels are a constellation of satellites deployed by ESA as part of the Copernicus program, which include high-resolution optical images from Sentinel-2A and 2B. The Sentinel-2s are sun-synchronous and multispectral instruments (MSI).  Sentinel band resolutions and wavelengths used in this study are shown in Table \ref{tab:sentinel-bands}. 

\begin{table}[h!]
    \centering
\caption{The used spectral bands of Sentinel-2.}
\label{tab:sentinel-bands}
\begin{tabular}{@{}lcc@{}}
%\toprule
Bands              & \begin{tabular}[c]{@{}c@{}}Resolution\\ m\end{tabular} & \begin{tabular}[c]{@{}c@{}}Wavelength\\ $\mu m$ \end{tabular} \\ 
%\midrule
B01 - Aerosols     & 60                                                     & 0.44                                                   \\
B02 - Blue        & 10                                                     & 0.50                                                   \\
B03 - Green       & 10                                                     & 0.56                                                     \\
B04 - Red         & 10                                                     & 0.66                                                   \\
B05 - Red edge 1  & 20                                                     & 0.70                                                   \\
B06 - Red edge 2  & 20                                                     & 0.74                                                   \\
B07 - Red edge 3   & 20                                                     & 0.78                                                   \\
B08 - NIR         & 10                                                     & 0.83                                                   \\
B09 - Water vapor & 60                                                     & 0.94                                                     \\
B10 - Cirrus      & 60                                                     & 1.37                                                  \\
B11 - SWIR 1      & 20                                                     & 1.61                                                  \\
B12 - SWIR 2      & 20                                                     & 2.20                                                  \\ 
%\bottomrule
\end{tabular}
\end{table}

To access and process the huge Sentinel-2 COPERNICUS/S2 image collection, we used the Google Earth Engine platform (GEE). The GEE \citep{gorelick2017google} is a cloud computing platform designed to store and process large datasets. The easily accessible and user-friendly front-end provides a convenient environment for interactive data and algorithm development. Google has archived all the Landsat and Sentinel image datasets and linked them to the cloud computing engine for open source use.  As well as providing the computational infrastructure and the image collections, the GEE API provides image pre-processing tools to resolve some common problems with the imagery such as cloud cover. The GEE Image Collections are stacks of raster images that can be processed as whole on single API calls, making it easy to process complex operations over a selection of images and their bands. This structure allows operations like filtering, mapping, reducing, compositing and iterating on large image sets. The results can be immediately visualized on screen or exported to a Google drive folder. The same API can be used from a Python Jupyter Notebook interface on Google Colaboratory. 

\begin{figure}[h]
\centering
\includegraphics[width=0.8\columnwidth]{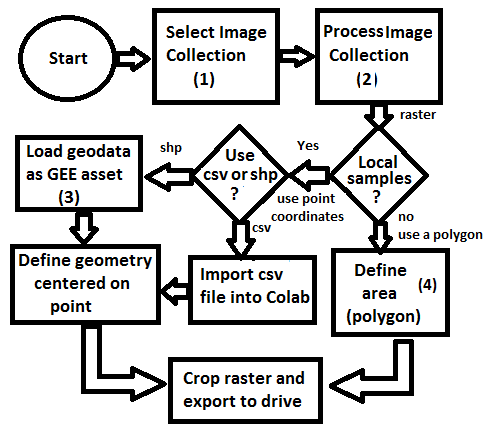}
\caption{Workflow for accessing and processing Sentinel 2 images.}
\label{fig:workflow}
\end{figure}

In order to acquire and prepare the image sets we followed the workflow presented in Figure \ref{fig:workflow}.  In the first step (1), the desired data collection is selected and then date and cloud cover filters are applied. To ensure acquisition of good quality images, a broad time-frame of two years, starting January 2018, was chosen. Using the metadata of the images, it was also possible to select only the images within this time-frame with less than 20 percent cloud cover. In the second step (2), the image collection is processed in order to obtain a single raster image. A pixel based function was applied to the image stack to remove pixels containing cloud or cloud shadow patterns evident in the QA60 bit band. Finally, to define a single raster image a pixel based median operation was applied to the image stack.  The median value at a location was the most common valid (i.e. none cloud) pixel value from all images at that location. This operation is based on the assumption that clouds are not in the same place permanently. 

\begin{figure}
\begin{subfigure}{0.48\textwidth}
\includegraphics[width=\linewidth]{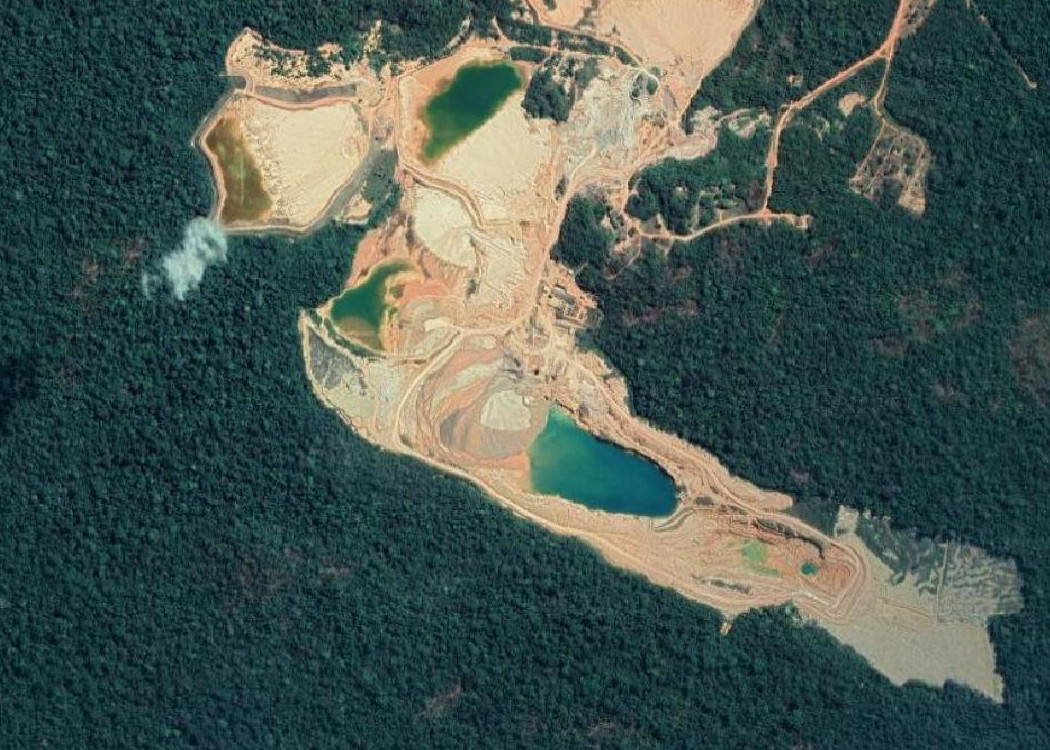}
\caption{High resolution Google Earth imagery showing mine and associated dam}
\label{fig:subfig1}
\end{subfigure}
\hspace*{\fill}
\begin{subfigure}{0.48\textwidth}
\includegraphics[width=\linewidth]{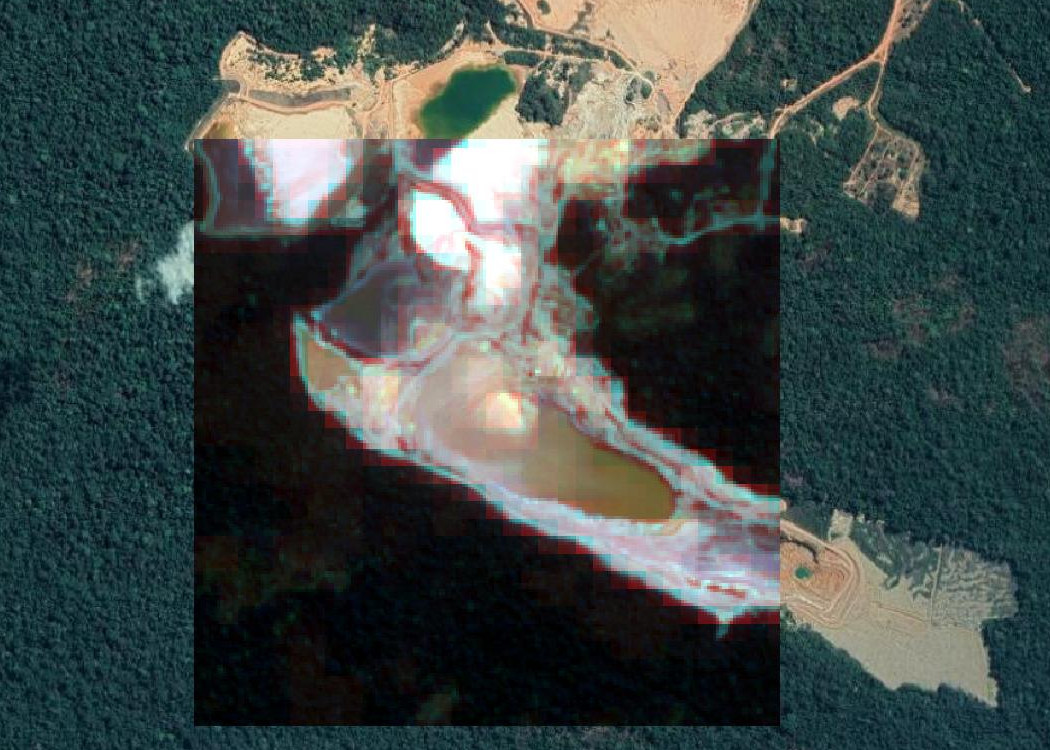}
\caption{Sentinel 2 imagery of mine and dam shown in sub-figure (a)}
\label{fig:subfig2}
\end{subfigure}
\caption[]{Example of Sentinel-2 image of dam and mine.}  \label{fig:sentinelsample}
\end{figure}

For input to our classification algorithm, the coordinates of the points of interest must be in tabular format, and made available as comma-separated values (csv) or Esri shapefile (shp). To facilitate the processing of shapefiles we import them into GEE as feature collections before loading them into Google Colab as arrays using GeoPandas \footnote{https://geopandas.org/} (i.e. workflow Step 3).  A small square image is extracted around each coordinate centred on the coordinate.  Each of these sub-images contains enough information to determine the presence, or not, of a mine or dam. Two sets of image crops were prepared. The first one had 2 x 2 km (0.018 x 0.018 degrees) image extracts, each comprising 201x201 pixels and 12 spectral bands. These small image extracts were used to train and validate the mine discovery classifier. The second set of cropped images, comprised only the point coordinates for which the mining ore was annotated, had 200 x 200 metres (0.0018 x 0.0018 degrees) image extracts, each comprising 21x21 pixels and 12 spectral bands.  This second dataset was used to train the environmental impact classification model. Figure \ref{fig:sentinelsample} shows an example of a mine and its tailings dam followed by the corresponding cropped image from the Sentinel-2 collection obtained using the workflow in Figure~\ref{fig:workflow}.

\subsection{Deep Learning}

Deep learning methods have obtained an unprecedented success in computer vision. High accuracy has been achieved on complex tasks such as the ImageNet Large Scale Visual Recognition Challenge (ILSVRC) to evaluate algorithms for object detection and image classification at large scale \citep{imagenet_cvpr09}. However, several aspects need to be taken into consideration when applying these methods in real life cases. State-of-the-art computer vision algorithms use thousands or sometimes millions of images with known annotations for training and evaluation on large scale problems. For real life cases this can be unfeasible due to the lack of examples or dubious interpretations of the object classes present in an image.  A level of acceptable automated classification is also task-specific in real life applications. In applications where a classification mistake has a high cost and may cause harm, inaccuracies inherent in machine learning methods may not be tolerated. In applications where automated classification methods are used only to support expert decisions, a small number of false positive or false negative classifications may be tolerated. In this research, which looks for indications of the presence of surface mines or tailings dams as well as of the type of associated environmental impact they have, it would perhaps be impracticable to obtain a highly accurate system given the limited amount of training examples and the numerous variations in shape, hue, areas and volumes of different mines and ore categories. Thus, we designate a complete analysis strategy, which aims to support the analyst in their discovery and classification of mines and dams in the most diverse regions of the planet without guaranteeing accuracy or performance levels. 

\subsubsection{The Convolutional Neural Network}

Advances in deep learning in computer vision have been strongly driven by Convolutional Neural Networks (CNNs) in recent years \citep{NIPS2012_4824}.  CNNs evolved from multilayer perceptrons, which are networks consisting of an input and an output layer, as well as multiple hidden layers, and use a gradient descent based optimization algorithm to update their parameters. A Neural Network is defined by its layer architecture and activation functions. A CNN typically comprises a series of convolutional layers, each performing local "filter" operations that detect features in the output of previous layers. The neural net learns the filters that in traditional computer vision algorithms are hand-engineered.  Other types of layers are also used in CNNs such as pooling, fully connected and regularization layers. Pooling layers combine the outputs of groups of neurons in one layer into a single neuron in the next layer in order to reduce data dimensionality. A fully connected layer connects all neurons in one layer to all neurons of the next layer. They usually appear after the convolutional layers so that the fully-connected layers capture complex relationships between high-level features extracted by the convolutional layers. CNNs may be prone to overfitting. To mitigate this undesired effect they are usually  supplied with regularization layers. One method to reduce overfitting is {\it dropout} where, at each training cycle, the outputs of some nodes are ignored with a given probability to prevent the nodes from co-adapting to the data. The activation function is responsible for transforming the weighted sum of a node's input into that node's output value. The activation function commonly used in CNNs is the Rectified Linear Unit (RELU). For a review of the main concepts related to deep learning please refer to \citep{goodfellow2016deep}.

\subsubsection{Visual Pattern Mining}

We explore the application of CNNs on two related environmental problems: the discovery of surface mines with their corresponding tailings dams in raw satellite imagery covering large areas (described in Section \ref{sec:scratch}) and the classification of the environmental impact of the discovered mines and dams (Section \ref{sec:oretype}).  These problems required some further adaptation of the CNN methodology which we describe next.

\begin{figure}
\begin{subfigure}{0.48\textwidth}
\includegraphics[width=\linewidth]{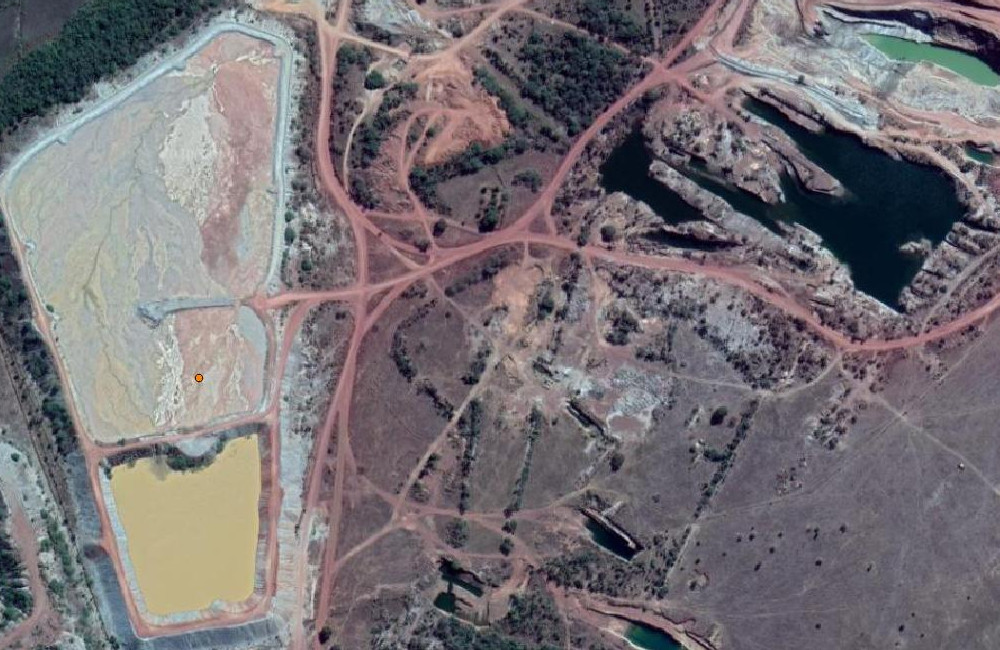}
\caption{Gold mine.}
\label{fig:subfig11}
\end{subfigure}
\hspace*{\fill}
\begin{subfigure}{0.48\textwidth}
\includegraphics[width=\linewidth]{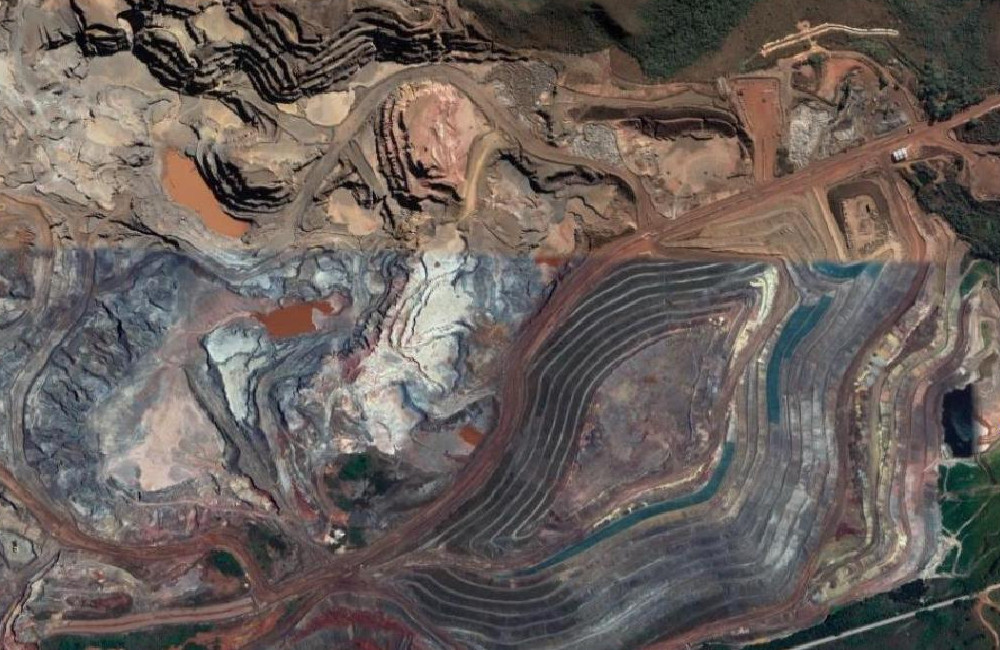}
\caption{Iron mine.}
\label{fig:subfig22}
\end{subfigure}
\caption[]{Visual patterns of mines (Google Earth).}  \label{fig:patterns}
\end{figure}

Computer vision problems are usually categorized as {\it image classification} when the goal is to classify one object within a whole image, as {\it object detection} when the goal is to classify and detect one or more objects within an image displaying their position with bounding boxes and {\it semantic segmentation} when the goal is to recognize the object class for each pixel of an image.  Our research problems do not fit completely into any of these categories. Mines and dams frequently do not have clear edges to delimit them in the imagery (see, for example,  Figure~\ref{fig:patterns}). Their sizes and shapes vary enormously. For these reasons they can hardly be considered "objects" to be detected, segmented or classified. An alternative machine learning approach to computer vision problems is Visual Pattern Mining \citep{DBLP:journals/corr/LiEZC17}.  In essence, VPM classifies parts of an object.  VPM is often implemented as a two phase approach. Individual parts of an object are classified in Phase 1 and then Phase 2 classifies the object by using spatial relationships of the parts identified in Phase 1. In Phase 2 the classification of visual patterns is obtained from the combination of multiple filters from the final convolutional layer of the Phase 1 CNN. Visual Pattern Mining is widely used for mid-level feature representations on image classification tasks and visual summarisation.  VPM code is freely available via the Python implementation PatternNet.

We adapted the pattern mining method for application to both mine and dam identification in satellite imagery as well as the classification of different types of construction and the corresponding environmental impact. Our adaptation uses Phase 1 of the VPM approach to identify parts of dams or mines. We do not distinguish the nature of the individual parts themselves, only that a sub-image of interest either contains or does not contain a part of a dam or mine. This allows us to fix the size of the CNN input image, and thus providing us with an efficient classifier that scales over large areas of interest.   The new algorithm was trained on small image samples, some of which contained a part of a dam or mine, annotated with the presence or not of a mine or tailings dam and the ore type if it was known.  We also implemented the pattern mining method as a fully convolutional network (FCN) for computational efficiency and scalability.  Since a FCN contains only convolution and pooling layers, with fully connected layers converted to convolution layers, it can be executed on large area images.  This is particularly useful for our research problems, where training needs to be done on small image patches, showing parts of a mine or a dam, but classification is performed on big images covering wide areas of Brazil. When applied to wide areas, output of the FCN is a map comprising clusters of class markings, each cluster marking a dam or mine. Our approach is described in more detail in Section \ref{sec:scratch}. 

The FCN algorithm was implemented using open-sourced computational resources which are freely available and easily accessible online.  CNN-based applications can be developed and deployed using open source libraries such as the Keras TensorFlow's high-level application programming interface (API), used for building and training deep learning models \citep{tensorflow}. It is useful for fast prototyping, state-of-the-art research, and production. The Keras TensorFlow API is available on Google Colaboratory (Colab), a free cloud service based on Jupyter Notebooks that offers robust Graphics Processing Unit (GPU) based computing at no cost \citep{Carneiro2018}. On Google Colab a developer has access to a flexible runtime fully configured for deep learning, avoiding all the workload required to prepare the infrastructure and environment that machine learning projects typically require. Large files can be uploaded, accessed and saved using Google Drive, greatly simplifying data input-output operations. The existence of such a platform both simplifies and reduces the cost of developing and running sophisticated machine learning methods, including deep learning, thus allowing its use for social good by NGOs and GOs in developing countries and not just by conventional academic and business organisations.

\subsubsection{Mine and tailings dam discovery} \label{sec:scratch}

 Our FCN architecture comprises five convolution layers, three pooling layers and a dropout layer with a 50\% dropout rate for regularization. The model contains 259,810 parameters, all trainable and distributed on the network layers as presented in Table \ref{tab:cnn-summary}. All convolution layers uses the rectified linear unit (RELU) activation function. The input training data is composed by 201x201x12 float arrays. They correspond to the 12 spectral bands of 2x2 km raw Sentinel-2 image samples. These images are normalised with each band having zero mean and unit standard deviation across the entire training data. Only the number of bands in the imagery is fixed (i.e.12) in the FCN input layer, the width and height of the input images are not specified to allow different sized images for training and classification. The output layer is a convolution with a 1x1 kernel and two filters, corresponding to two target classes: \emph{mine} and \emph{not mine}. The \emph{mine} class corresponds to scenes containing surface mines and/or tailings dams.

\begin{table}[h]
\centering
\caption{Mine and dam discovery model summary}
\label{tab:cnn-summary}
\begin{tabular}{c c c c c}
\hline
\textbf{Layer} & \textbf{Type} & \textbf{Filters\#} & \textbf{Kernel} & \textbf{Param\#} \\

\hline
input & InputLayer  & &   & 0     \\    
conv2d\_1  & Convolution & 32 & 3x3 & 3488      \\
maxpooling1 & MaxPooling & & 3x3  &  0         \\
conv2d\_2 & Convolution & 64 & 3x3 & 18496     \\
maxpooling2 & MaxPooling & & 3x3 &  0     \\    
conv2d\_3 & Convolution & 64 & 3x3 & 36928     \\
maxpooling3 & MaxPooling & & 3x3 &   0      \\   
conv2d\_4 & Convolution & 64 & 7x7 & 200768   \\
dropout & Dropout 50\% & & & 0         \\
conv2d\_5 & Convolution & 2 & 1x1 & 130       \\\hline
\end{tabular}
\end{table}

\begin{figure}
\centering
    \includegraphics[width=0.7\columnwidth]{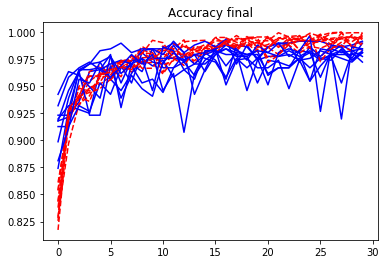}
\caption{Evolution of accuracy for mines and dams detection over 30 training epochs and 10-fold cross validations. Red lines indicate training accuracy and blue lines validation accuracy.}
\label{fig:training}
\end{figure}

Our FCN architecture is rather conventional and the implementation of its training, validation and testing procedures is straightforward on the Keras TensorFlow API. The fact that the input data consists of raster images, stored as TIFF (Tagged Image File Format) files, requires proper load and translation steps in order to prepare the float arrays (or tensors) required as input to the CNN. We used the Geospatial Data Abstraction Library (GDAL)\footnote{https://gdal.org/} to implement these steps. 2860 201x201x12 images were used to train and test the FCN. The model takes approximately 100 seconds to train on Colab GPUs. Figure \ref{fig:training} shows the iterative evolution of the model accuracy on ten re-sampling, training and validation procedures (k-fold cross validation). We obtained an accuracy on the training data of $97.39\pm 0.03\%$ after 30 epochs. In each re-sampling step the FCN was validated on a separate dataset containing 572 201x201x12 images. The average validation accuracy was $96.85 \pm 0.03\%$, again after 30 epochs. The average F1 score was 0.9689 and the average confusion matrix was :

\[
  \begin{bmatrix}
   & \text{mine} & \text{not\ mine}\\ 
    \text{mine} & 272.6 & 5.1\\
    \text{not\ mine} &  5.0 & 289.3
  \end{bmatrix}
\]

\subsubsection{Environmental impact classification} \label{sec:oretype}
 Various principles and processes are utilised within mining that determine the impact they have on the environment \citep{norad}:

\begin{itemize}
\item Low environmental impact: extraction and any processing of sand, gravel or rock where the environmental impact is usually restricted to dust or noise. It is usually related to extraction for construction purposes or as raw materials for industry. The processing does not go beyond crushing or screening of sand and gravel. 
\item High environmental impact: extraction of minerals and ores in facilities which produce large amounts of waste consisting of finely crushed rock (tailings) and discharges of polluted waste water. This category comprises mining sites where the valuable substances in the ores must undergo processing and where large amounts of tailings must be deposited. Within this category, a sector of great environmental importance is the mining of sulphide ores containing copper, zinc, lead or nickel, often in combination with pyrite. These ores are the most common ores on a global scale. The environmental impact of this type of mining can last a long time after the end of activities\footnote{An exception to this simplified classification of environmental impact concerns gold mines. Although its process is similar to that used in the extraction of gravel, based on washing, crushing and screening, the environmental impact can be great due to the damage it can cause to water courses and the use of products that are harmful to the environment such as mercury. However, in low resolution satellite images it is not possible to distinguish a gold mine from a simple exploration of sand or gravel.}
\end{itemize}

In order to provide an indication of the potential environmental impact of a discovered mine or tailings dam we explored again the usefulness of deep learning based classifiers.
Using the Brazilian officially declared mines and tailings dams databases, containing the ore type declared by their owners, we chose 1,275 representative point coordinates and acquired Sentinel 2 image patches, using the protocol described in Section \ref{sec:data}. Each image patch has 21x21 pixels and 12 spectral bands.  The classes considered in this second predictive model, their related materials and the number of image samples used as training dataset are:
\begin{enumerate}
\item high impact waste: lead, iron, niobium, zinc, copper, phosphate, limestone, nickel, schist, manganese and bauxite; 425 samples.
\item low impact waste: sand, gravel, clay, stones ; 425 samples.
\item no ore: soil, rocks and vegetation; 425 samples.
\end{enumerate}

This second model used a conventional CNN network architecture, consisting of two convolution layers followed by a max pooling layer each and two dense layers. Due to the smaller 21x21 training image patches the convolution layers used a reduced kernel of size 2x2. The max pooling layers also used a 2x2 kernel. The first dense layer has 1024 elements. The second has 3 elements, corresponding to the output classes, and uses a softmax activation.  The architecture, comprising 2,373,219 parameters, all trainable, is presented in Table \ref{tab:cnn2-summary}. 

\begin{table}[h]
\centering
\caption{Environmental impact model summary.}
\label{tab:cnn2-summary}
\begin{tabular}{c c c c c}
\hline
\textbf{Layer} & \textbf{Type} & \textbf{Filters\#} & \textbf{Kernel} & \textbf{Param\#} \\

\hline
input & InputLayer  & &   & 0     \\    
conv2d\_1  & Convolution & 32 & 2x2 & 1568      \\
maxpooling1 & MaxPooling & & 2x2  &  0         \\
conv2d\_2 & Convolution & 64 & 2x2 & 8256     \\
maxpooling2 & MaxPooling & & 2x2 &  0     \\    
flatten & Flatten & &  &   0      \\   
dense\_1 & Dense & 1024 &  & 2360320    \\
dropout & Dropout 50\% & & & 0         \\
dense\_2 & Dense & 3 &  & 3075      \\
\\\hline
\end{tabular}
\end{table}

\begin{figure}
\centering
\includegraphics[width=0.7\columnwidth]{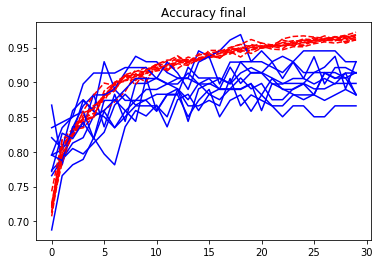}
\caption{Evolution of accuracy for environmental impact prediction over 30 training epochs and 10-fold cross validations. Red lines indicate training accuracy and blue lines validation accuracy.}
\label{fig:oretypetraining}
\end{figure}

Training the network  over 30 epochs takes on average 27 seconds on Colab GPUs. Figure \ref{fig:oretypetraining} shows the iterative evolution of the model accuracy. Using cross validation accuracy attained on average $92.70\pm 0.05 \%$. The average validation accuracy was $96.50\pm 0.003 \%$. The average confusion matrix was :

\[
  \begin{bmatrix}
        & \text{high\ impact} & \text{low\ impact} & \text{no\ ore} \\ 
    \text{high\ impact} & 40.3 &   1.9 &  0.3 \\
    \text{low\ impact} &  1.6 & 36. &  4.9 \\
    \text{no\ ore} & 0.9 & 3. & 38.6 \\
  \end{bmatrix}
\]

\subsubsection{Wide-Area maps of Dams and Mines} \label{sec:approach}

Google Earth Engine (GEE) provides processed image mosaics for large spatial areas.~\footnote{Our paper is accompanied by open-source Python scripts for Sentinel 2 imagery processing on GEE.} over which we wish to search for dams and mines. GEE breaks the processed mosaics into a number of GeoTIFF files that can be saved on Google Drive. Each file covers an area up to 92x92 kilometers (9216x9216 pixels and 12 bands). The volume of those files, around 3GB,  when used as input for FCN classification, exceeds the GPU memory available on Colab. To deal with this limitation we split each image into four near equally sized patches of approximately 46x46 kilometers (2,116 square kilometers) each. The FCN output for each patch is a 164x164 grid of probability distributions, representing the probability of each class on a certain region of the input image patch. 

For each grid point predicted as belonging to the \emph{mine} class we also classify its environmental impact (i.e. low, high or no ore). The environmental impact CNN uses as input a patch with 21x21 pixels (corresponding to an area of 210x210 meters) from the same mosaic taken around the location of the discovered mine. The CNN output is a probability distribution, representing the probabilities of each environmental impact class on that location.  If the \emph{no ore} class has the highest probability the predicted mine spot is considered a false positive and discarded. To determine the geographic coordinates of the detected mines from the FCN output map we used Rasterio\footnote{https://rasterio.readthedocs.io/en/latest/}, a Python API based on Numpy N-dimensional arrays and GeoJSON, to compute the coordinates of discovered targets. 

\begin{figure}
\centering
\includegraphics[width=0.8\columnwidth]{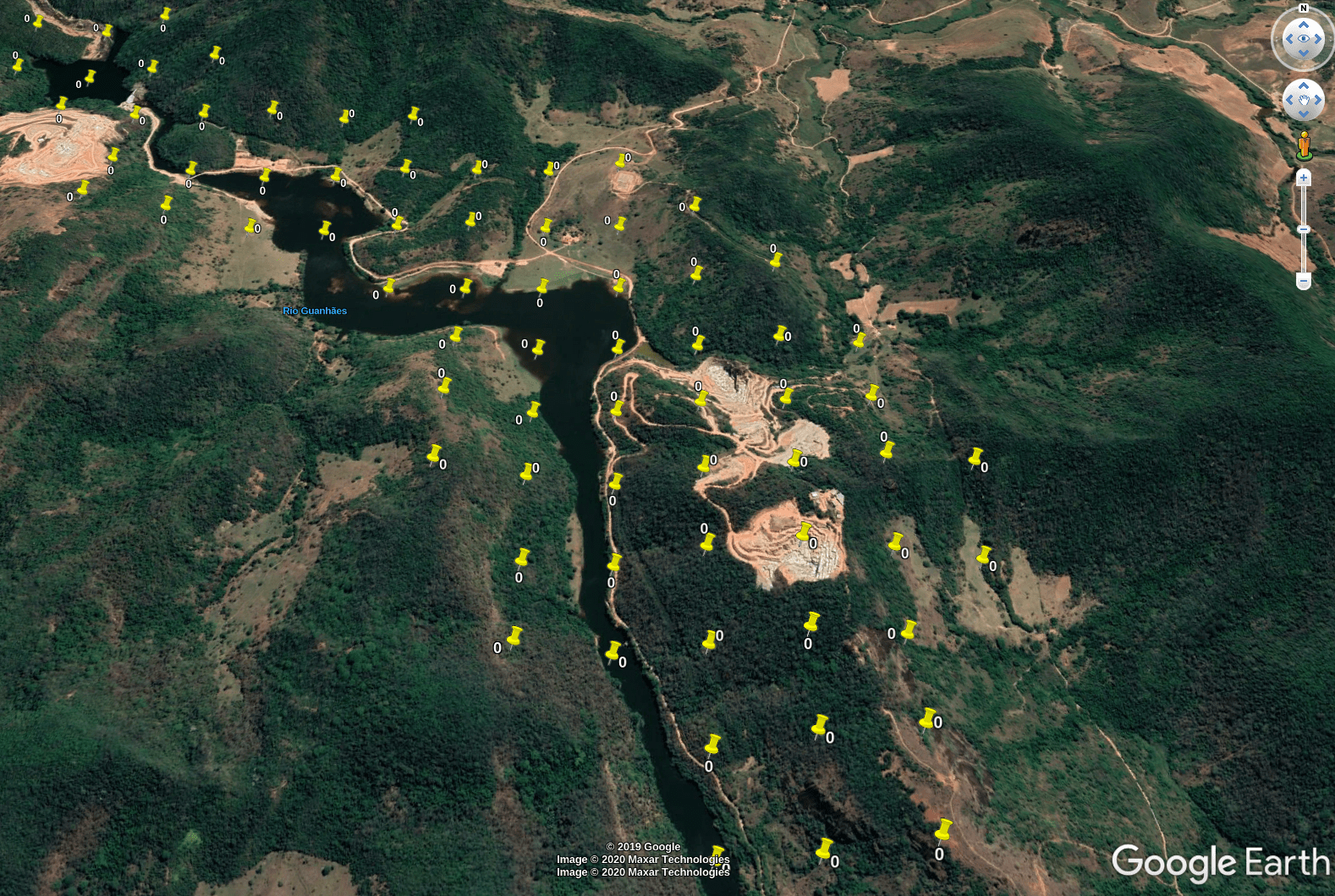}
\caption{Mines and tailings dams indicated by yellow marks on Google Earth.}
\label{fig:googleearthplot}
\end{figure}

The locations of the mines and tailings dams are exported as georeferenced delimited text files (csv and kml) which can be scrutinized using GIS software or Google Earth. We suggest an analyst visually inspect the CNN discovered mines and tailings dams against the raw satellite imagery or against a base map, like Google Earth's, to validate the discovered mines or tailings dams. Figure \ref{fig:googleearthplot} shows the dams and mines plotted on a Google Earth base map. 

\section{Experiments and Discussion} \label{sec:results}
\begin{figure}
\centering
\includegraphics[width=0.6\columnwidth]{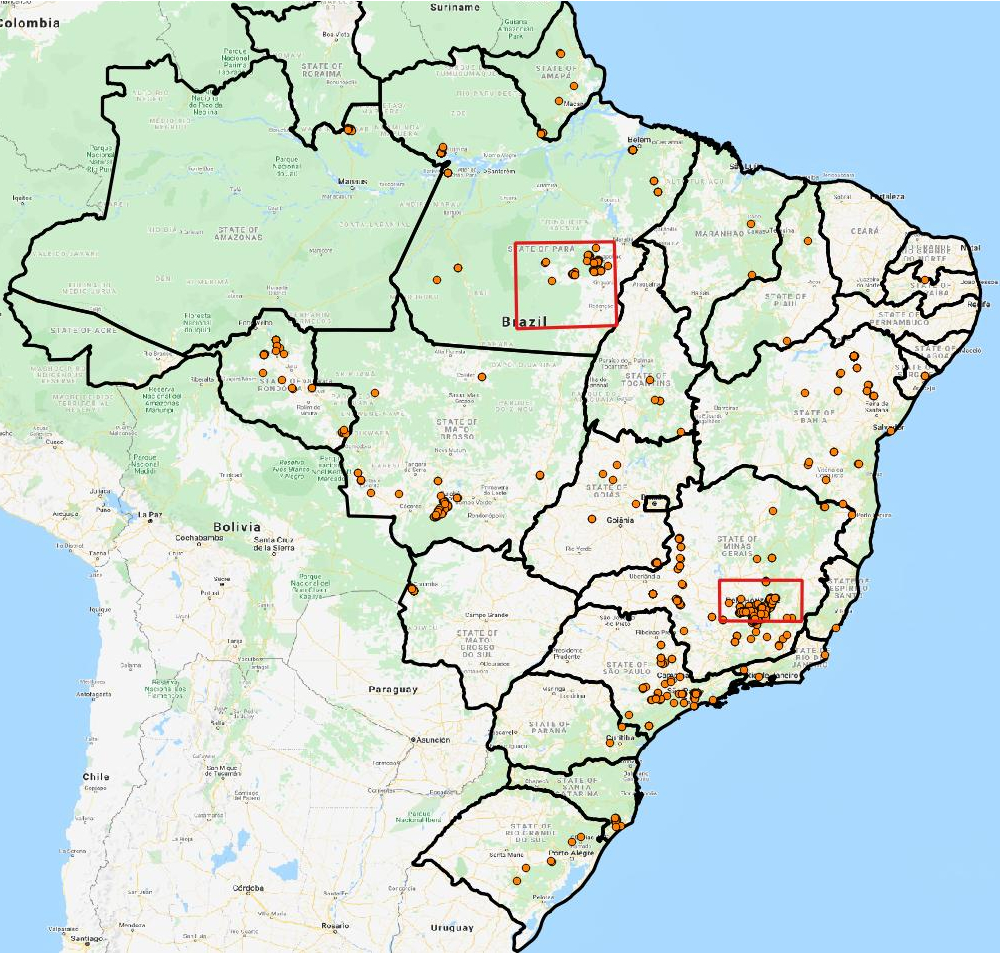}
\caption{Areas of interest: Minas Gerais (south) and Para (north).  Red dots are registered dams.}
\label{fig:areasanalisadas}
\end{figure}

We applied our approach to two large areas on the Brazilian territory searching for mines and dams and classifying their potential environmental impact. These areas are indicated in Figure~\ref{fig:areasanalisadas}. The first area covering 67,110 square-kilometres is located in the state of Minas Gerais. It is one of the areas with the highest concentration of mines and tailings dams in Brazil and includes the cities of Brumadinho and Mariana, where the cited tragic dam collapses occurred. The second area covers 170,267 square-kilometres and it is located in the state of Para where protected indigenous land and environment conservation units struggle against illegal mining and deforestation activities. It includes the Kayapo's protected land depicted in Figure \ref{fig:kayapo}.  Both areas are represented by red squares in Figure~\ref{fig:areasanalisadas} and the coordinates of the officially registered tailings dams from the ANA database are shown by orange dots. Raster images from both areas were acquired and processed using the workflow in Figure~\ref{fig:workflow}. In the "local samples" decision we follow the "no" path where the areas are defined directly as geo-polygons on step (4) before crop and export.

\begin{figure}
\centering
    \includegraphics[width=0.8\columnwidth]{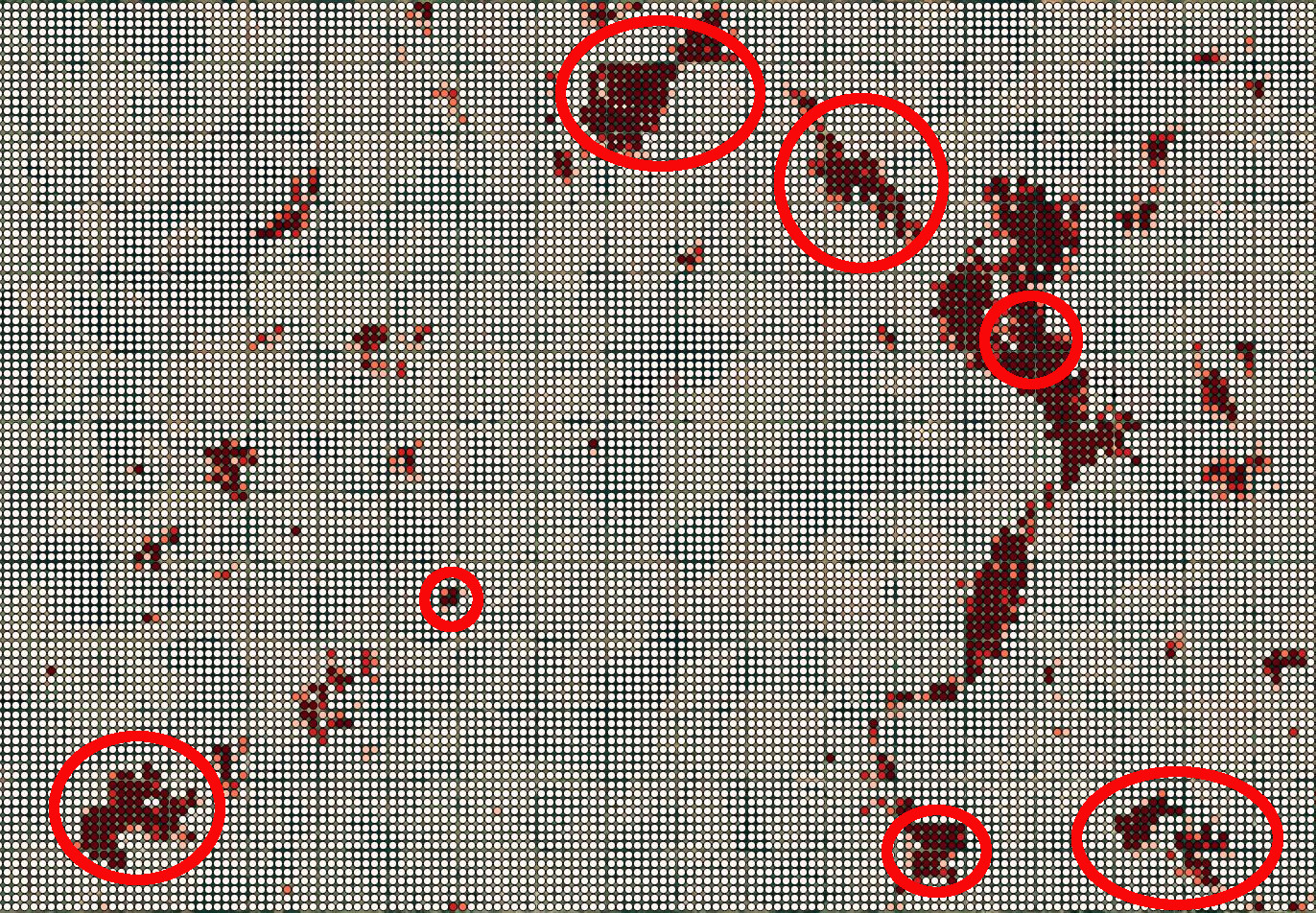}
\caption{Predicted mines in Minas Gerais area in Brazil.}
\label{fig:predictiongrid}
\end{figure}

\begin{figure}
\begin{subfigure}{0.48\textwidth}
\includegraphics[width=\linewidth]{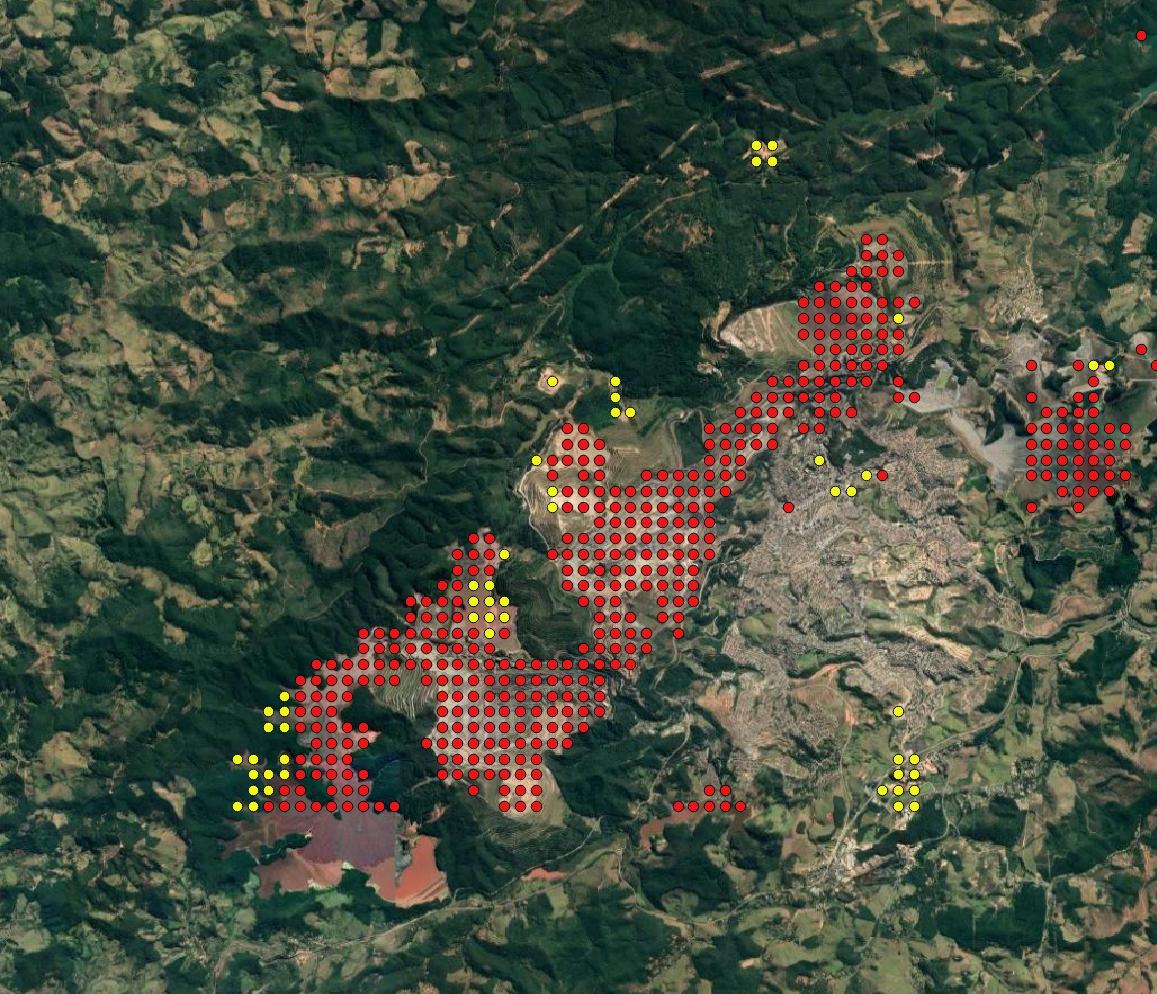}
\caption{Large mine: red indicating presence of mine with potential high  environmental impact}
\label{fig:mine1}
\end{subfigure}
\hspace*{\fill}
\begin{subfigure}{0.48\textwidth}
\includegraphics[width=\linewidth]{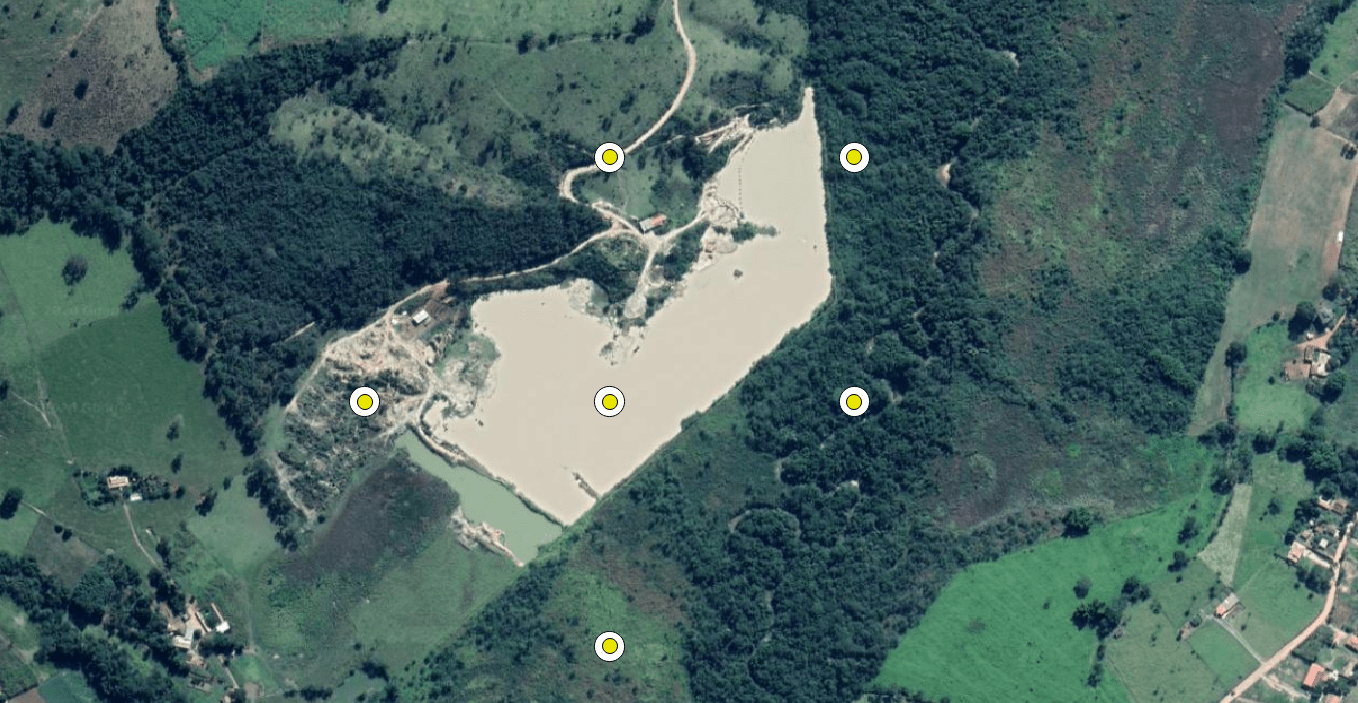}
\caption{Small mine: yellow indicating presence of mine with potential low environmental impact}
\label{fig:mine2}
\end{subfigure}
\caption[]{Sample true positives by the discovery model.}  \label{fig:truepositives}
\end{figure}

The FCN model was run on the 237,000 square kilometers of satellite imagery of Brazil and was able to identify all visible mines on those areas, even small ones of less than one square kilometer in extent. However, a considerable number of false positives were also present. An example of the typical output grid obtained using the discovery model is shown in Figure \ref{fig:predictiongrid}. This grid covers 1,120 square kilometers in Minas Gerais, Brazil. Dark areas show predicted mines. Red circles indicate where the predictions were confirmed by visual inspection. All other dark areas are false positives. Two true positives are shown in Figure \ref{fig:truepositives}. Figure \ref{fig:mine1} shows a large mine covering 12 square kilometers and Figure\ref{fig:mine2}, a smaller one with 0.7 square kilometers. 

\begin{figure}
\begin{subfigure}{0.48\textwidth}
\includegraphics[width=\linewidth]{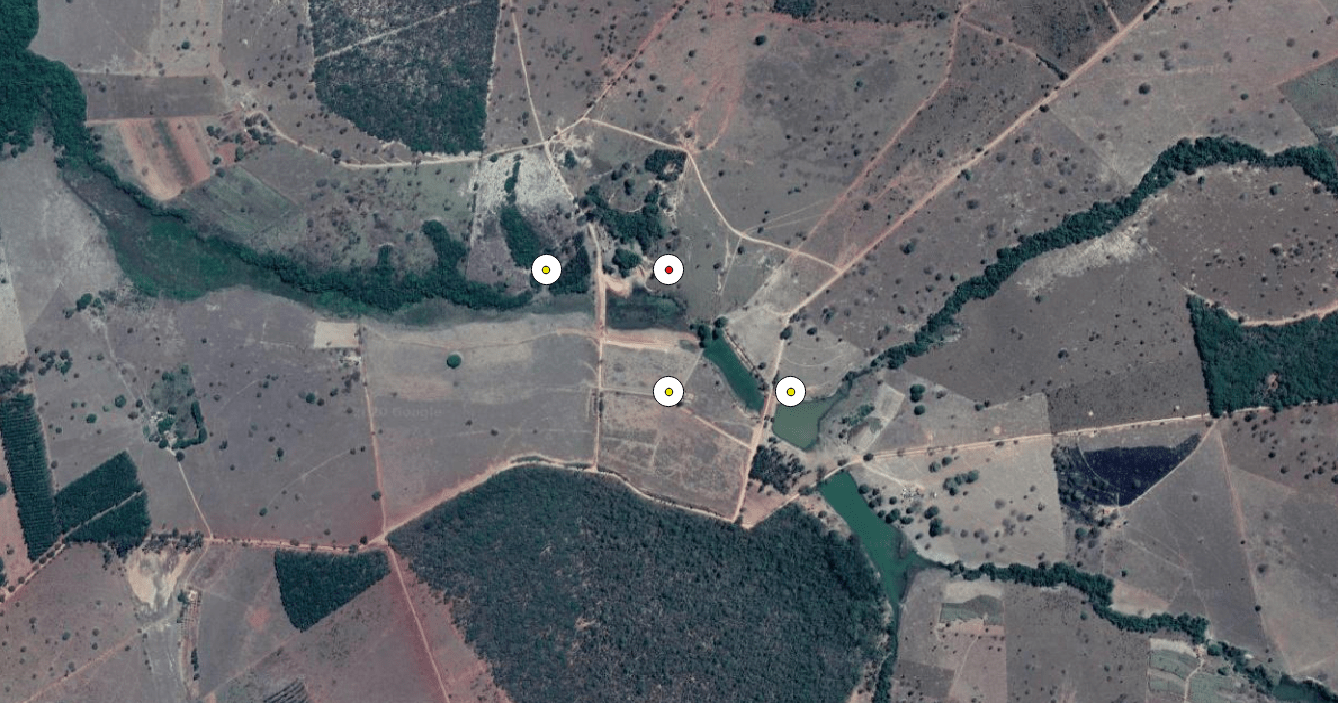}
\caption{Irrigation dam}
\label{fig:irrigationdam}
\end{subfigure}
\hspace*{\fill}
\begin{subfigure}{0.48\textwidth}
\includegraphics[width=\linewidth]{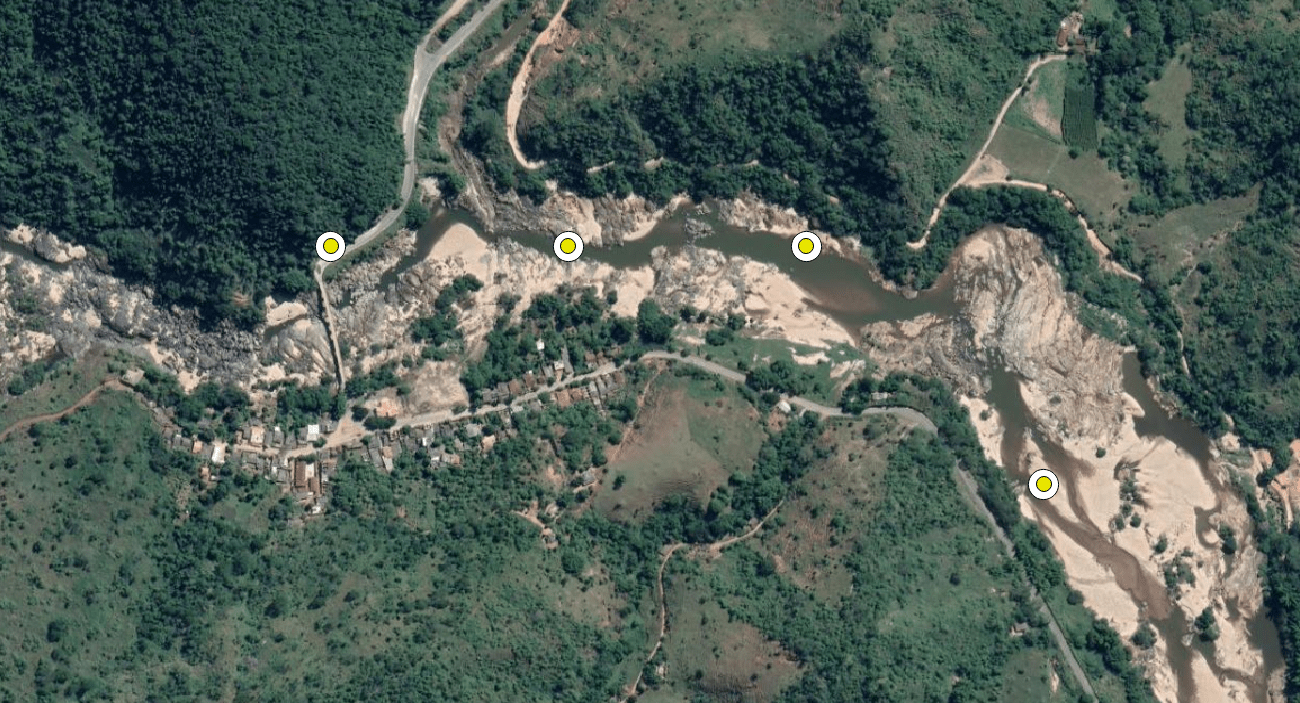}
\caption{River bank}
\label{fig:riverbank}
\end{subfigure}
\caption[]{Sample false positives by the discovery model.}  \label{fig:falsepositives}
\end{figure}

False positives occur in some very specific situations. Figure \ref{fig:falsepositives} shows two common situations, irrigation dams, depicted in Figure \ref{fig:irrigationdam}, and river banks shown in Figure \ref{fig:riverbank}. Feeding these false positives back into the training data and retraining the network would reduce the false positive rate. However, as the false positive rate was acceptable it was our choice to  correct these maps manually. 

\begin{figure}
\begin{subfigure}{0.48\textwidth}
\includegraphics[width=\linewidth]{images/exploracaoargila-min.png}
\caption{Low environmental impact: clay exploration}
\label{fig:lowimpact}
\end{subfigure}
\hspace*{\fill}
\begin{subfigure}{0.48\textwidth}
\includegraphics[width=\linewidth]{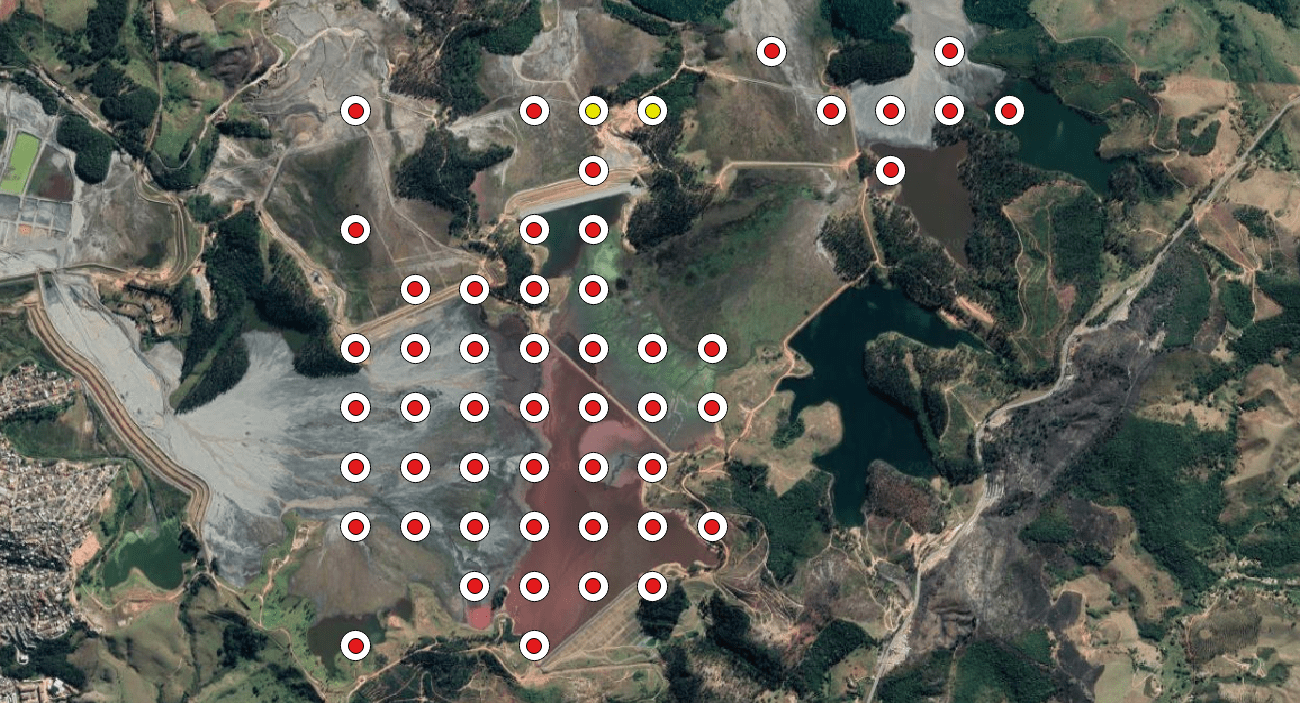}
\caption{High environmental impact: iron mine tailings dame}
\label{fig:highimpact}
\end{subfigure}
\caption{Sample of classified discovered mines. Red dots indicate high environmental impact mining and yellow low environmental impact mining.} \label{fig:impactclass}
\end{figure}

\begin{figure}
\begin{subfigure}{0.48\textwidth}
\includegraphics[width=\linewidth]{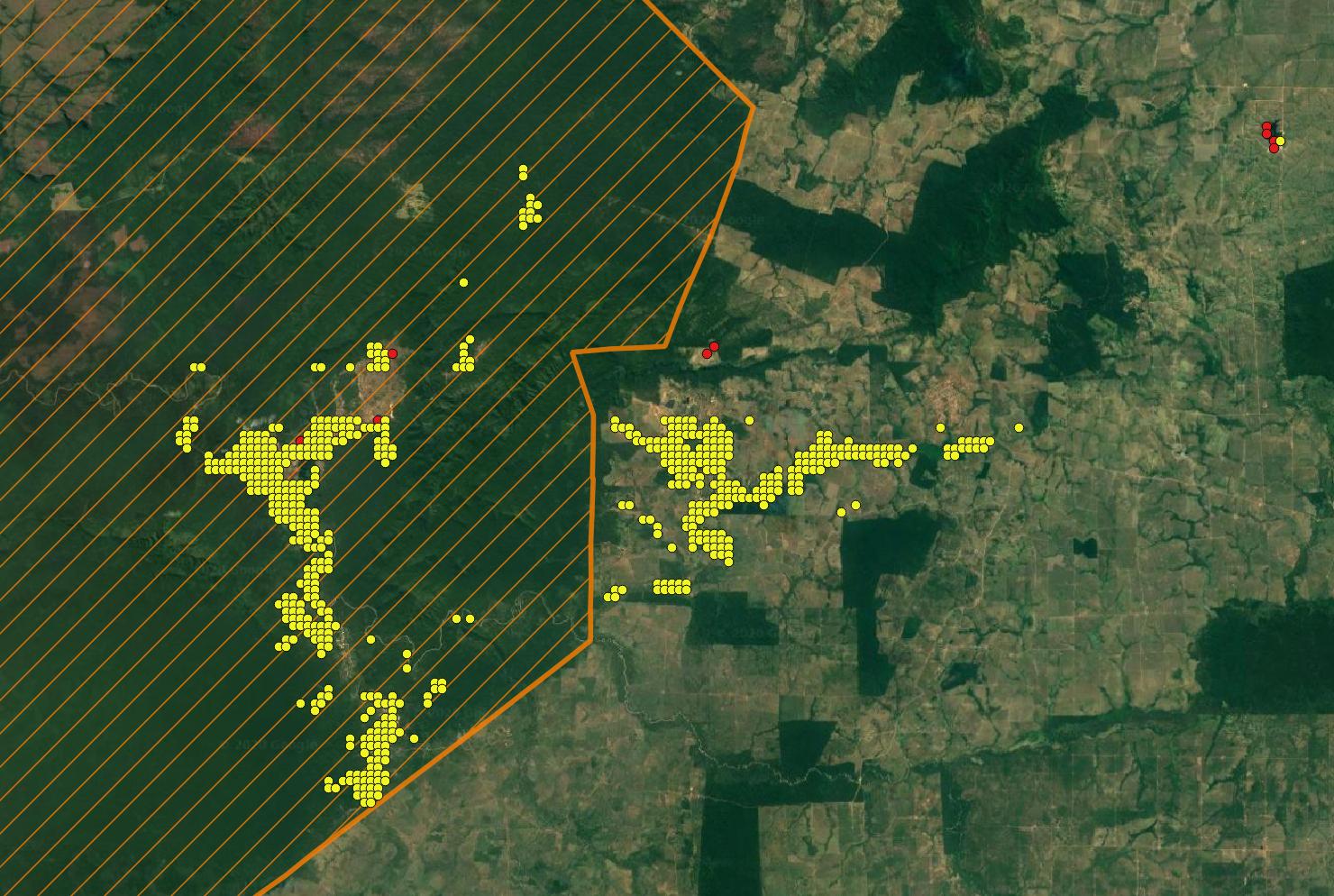}
\caption{Discovered spots on the Kayapo's protected land.}
\end{subfigure}
\hspace*{\fill}
\begin{subfigure}{0.48\textwidth}
\includegraphics[width=\linewidth]{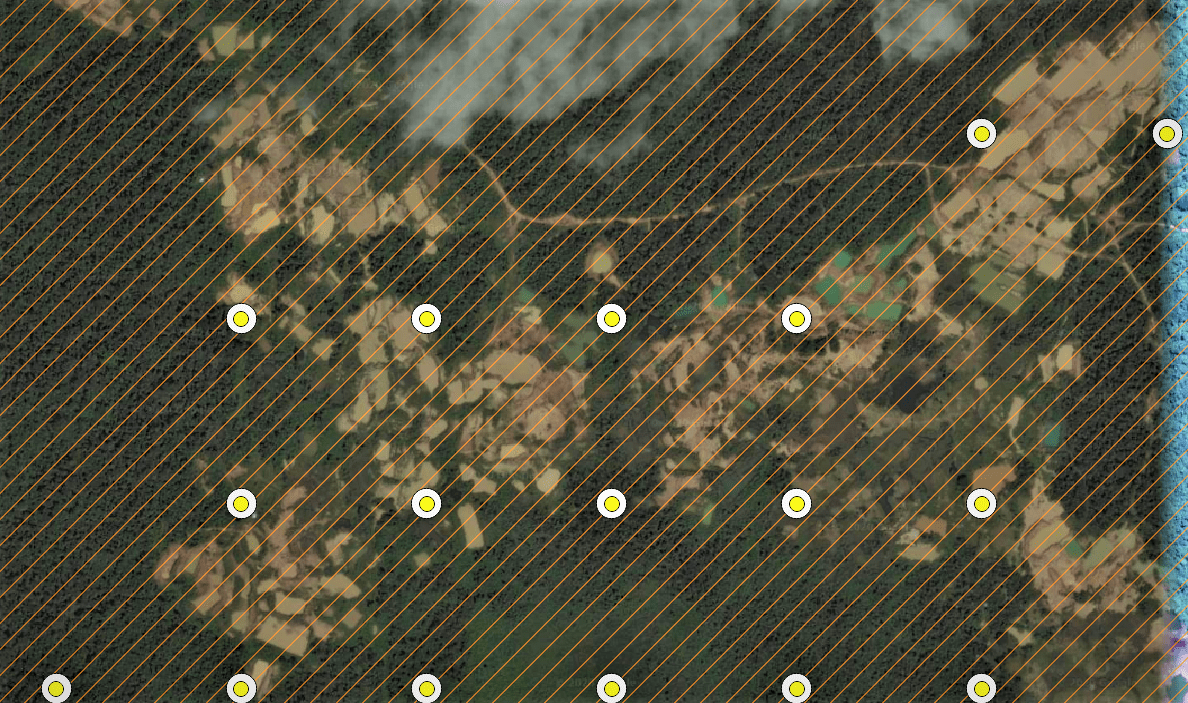}
\caption{Detail of a discovered mining spot on the Kayapo's protected land.}
\end{subfigure}
\caption{Mining identified by CNN in Kayapo's protected land. Yellow dots indicate the discovered mines. The orange line pattern fill indicates the Kayapo's land.} \label{fig:kayapomines}
\end{figure}

\begin{figure}
\centering
\includegraphics[width=.8\columnwidth]{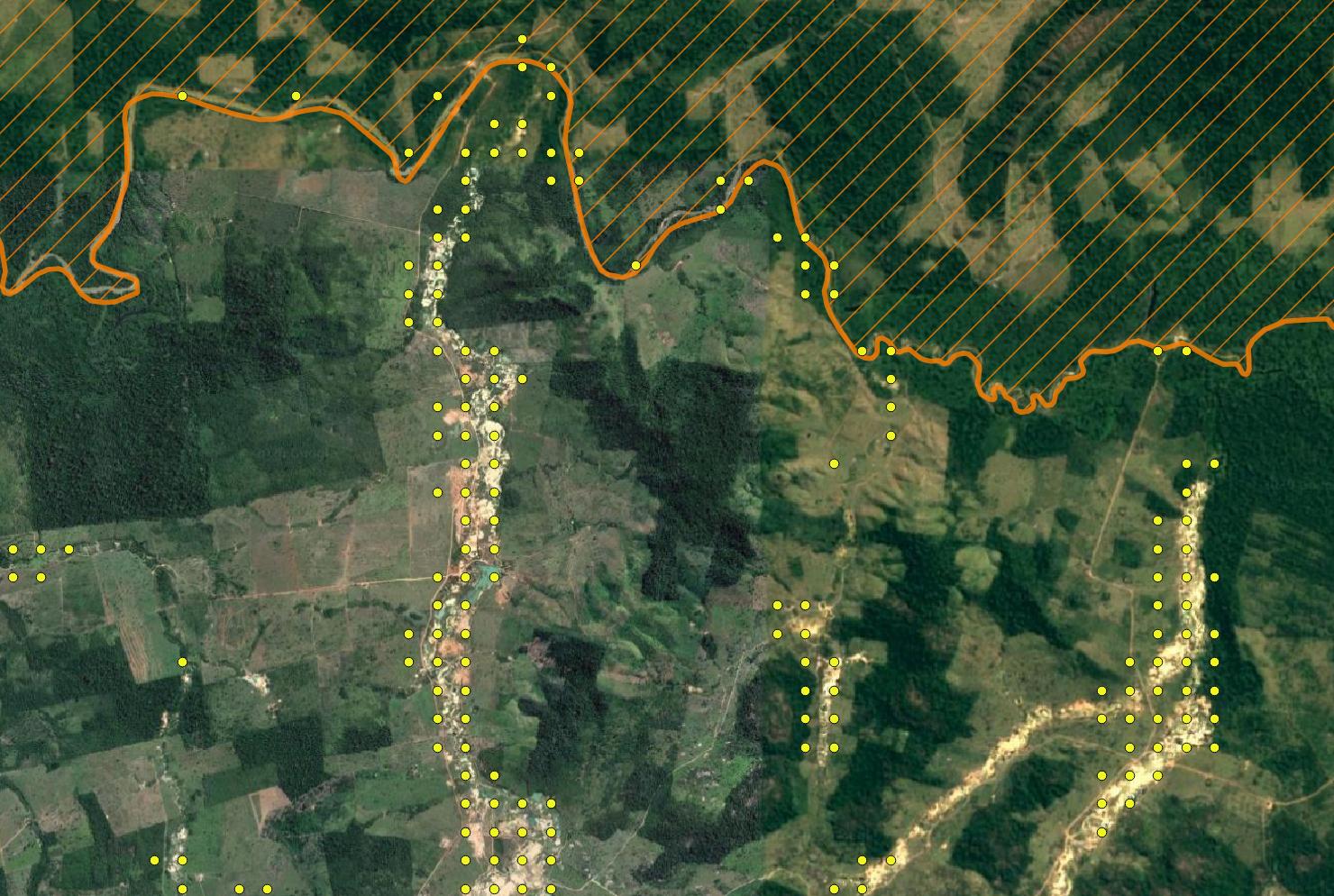}
\caption{Mining identified by CNN in Parakan's protected land. Yellow dots indicate the discovered mines. The orange line pattern fill indicates the Parakan's land.}
\label{fig:Parakanland}
\end{figure}

Visual inspection of the positive predictions for the area in Minas Gerais confirmed discovery of 154 clusters having mines and/or tailings dams. 63 of those clusters are newly discovered mines, not indicated on the official databases of the Brazilian government and not having an official mining concession. Visual inspection for the second area in the state of Para also confirmed a number of relevant positive predictions. 200 large unregistered gold mining areas were identified, many of them inside and around protected indigenous land, as can be seen in Figures \ref{fig:kayapomines} and \ref{fig:Parakanland}\footnote{Part of the discovered mines are in areas where a mining authorization was requested but the concession had not been carried out, which makes them illegal mines. }. Large iron mines were also detected inside two different environment conservation units. 

Using the mine detection model to find potential mines, and then applying the environment risk model to the mine images enabled the classification of the environment as shown in, for example, Figure \ref{fig:impactclass}, where two previously known mining sites where identified and correctly classified according to their environment impact. Figure \ref{fig:lowimpact} shows a sand mining site and Figure \ref{fig:highimpact} shows a high impact tailings dam in an iron mine.

\begin{figure}
\begin{subfigure}{0.48\textwidth}
\includegraphics[width=\linewidth]{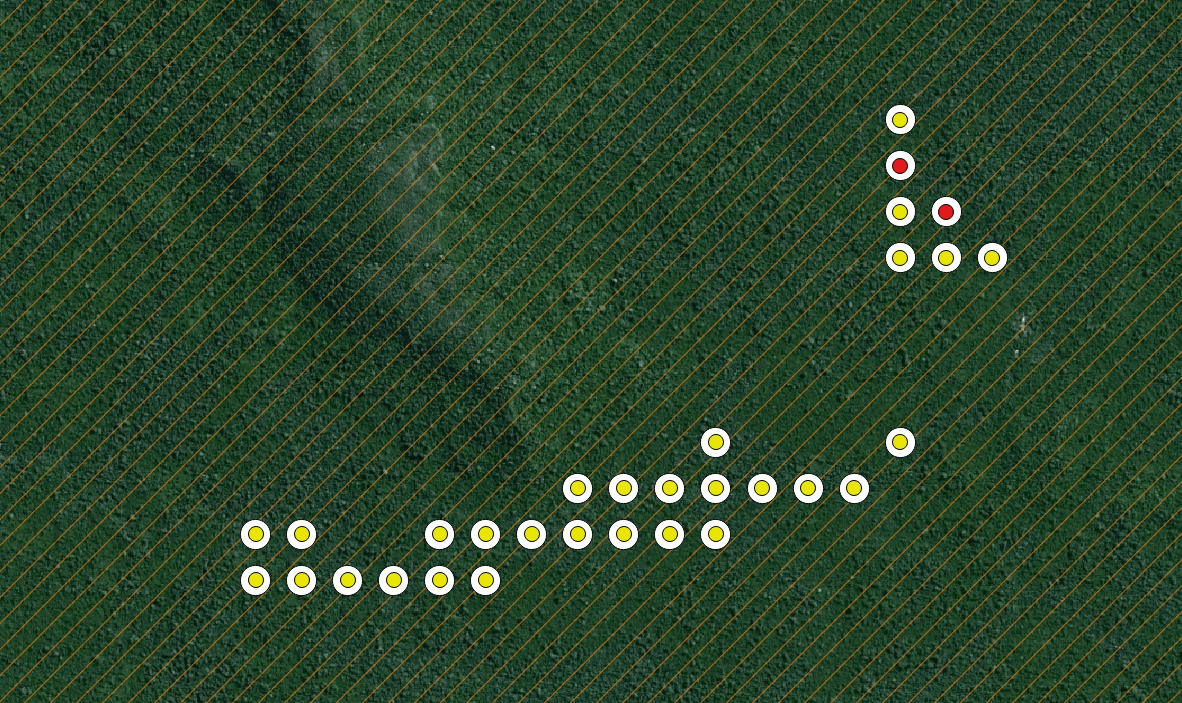}
\caption{True positives on the Kayapo's protected land not visible on Google Earth.}
\end{subfigure}
\hspace*{\fill}
\begin{subfigure}{0.48\textwidth}
\includegraphics[width=\linewidth]{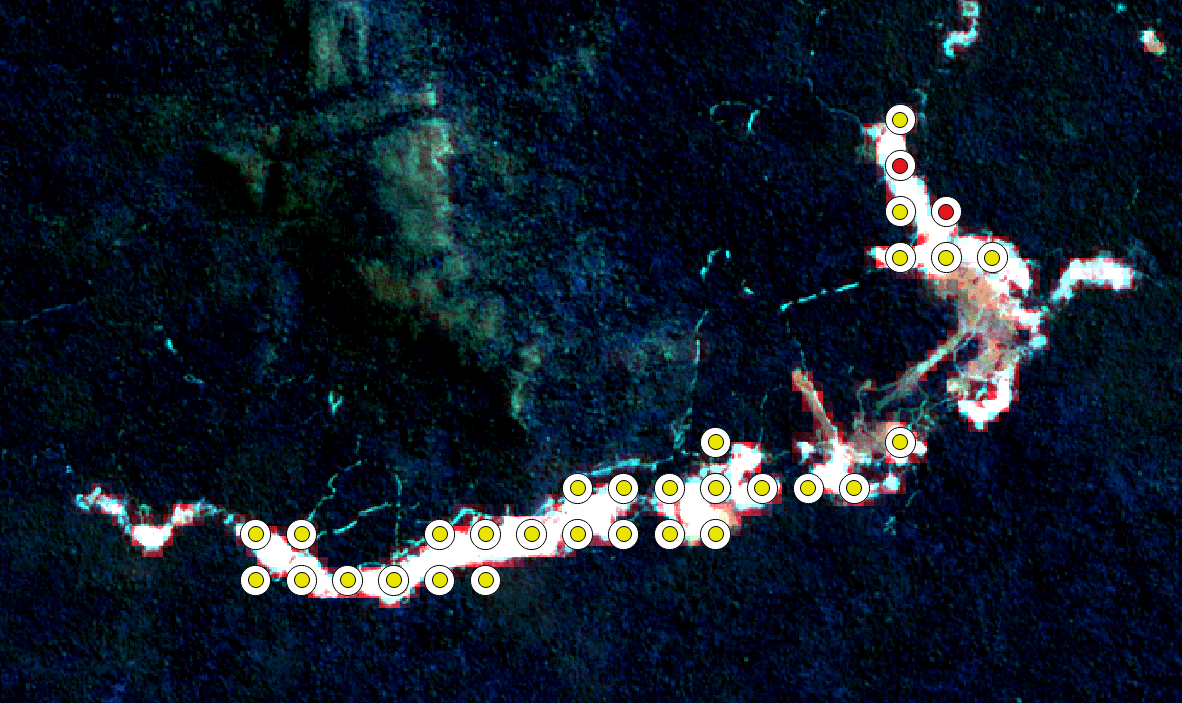}
\caption{True positives on Kayapo's protected land visible on Sentinel 2 images.}
\end{subfigure}
\caption{True positives not visible on Google Earth.} \label{fig:kayapomines2}
\end{figure}

We have demonstrated that mines can be detected over a wide area using free low resolution satellite imagery and automation. Although the majority of mines and dams are visible in images of good resolution, and can therefore be identified by a specialist without the aid of an algorithm, it is unlikely that this specialist will be able to in a cost effective way in very large areas. What's more, she will need high-resolution imagery which is either paid for or usually out of date when it becomes freely available. As it can be seen in Figure \ref{fig:kayapomines2}, some mines discovered at the Kayapo's reserve cannot be seen on Google Earth's outdated imagery. 

Despite the proliferation of deep learning toolboxes and APIs, applying models to new data is seldom straightforward \citep{Henderson2017DeepRL}. In our work, we took special care to allow a user to adapt our algorithm, models and data processing approaches to their own needs. Scripts and image datasets are provided, as well as models already trained and ready to be used on new imagery. Our models were trained with images from mines and tailings dams in Brazil, which certainly created a bias given the local characteristics of the soil and vegetation and the mining processes used there. The user can choose between applying the models as they are on their own data, retrain the CNNs from scratch using new imagery of mines and tailings dams, or improve the models using additional training with new images (i.e. transfer learning).

\section{Conclusion}  \label{sec:conclusions}
%\todo{The conclusion goes here.}

This work had an emphasis on using deep learning methods to solve a real large scale problem. Training and testing used small image patches, as usual. However, unlike conventional computer vision projects, the discovery of mines and dams was done on large areas. Fully convolutional deep neural networks were used in an innovative way to predict the presence of specific patterns on regions of the images. Mines and tailings dams can be discovered in large areas and their environment impact can be estimated. The case studies confirmed the usefulness of the approach, discovering 263 unregistered mining sites. Reproducibility was also one of our priorities. All design decisions maintained this practical and applied view of the use of free imagery, free cloud-based high performance computational infrastructure, free software APIs specialized on large scale image processing (GEE) and deep learning (TensorFlow).

The authors would like to thank FAPDF-Brazil for their financial support to this research project, Google X and the EPSRC for generous financial gifts and the LLoyd's Register Foundation for funding through the Alan Turing Institute Data Centric Engineering programme.

\bibliographystyle{elsarticle-num-names}
\bibliography{references.bib}






% \end{thebibliography}

\end{document}